\documentclass[10pt,twocolumn,letterpaper]{article}

\usepackage{cvpr}
\usepackage{times}
\usepackage{epsfig}
\usepackage{graphicx}
\usepackage{amsmath}
\usepackage{amssymb}
\usepackage{booktabs}       
\usepackage{commath}
\usepackage{mathtools}
\usepackage{subcaption}
\usepackage{algorithm}
\usepackage{algorithmic}
\usepackage{pdfpages}

\usepackage[breaklinks=true,bookmarks=false]{hyperref}

\cvprfinalcopy 

\def\cvprPaperID{****} 

\begin{document}
\title{Task-Adaptive Feature Transformer for Few-Shot Segmentation}

\author{Jun Seo\\
	KAIST\\
	{\tt\small tjwns0630@kaist.ac.kr}
	\and
	Young-Hyun Park\\
	KAIST\\
	{\tt\small dnffkf369@kaist.ac.kr}
	\and 
	Sung-Whan Yoon\\
	UNIST\\
	{\tt\small shyoon8@unist.ac.kr}
	\and
	Jaekyun Moon\\
	KAIST\\
	{\tt\small jmoon@kaist.edu}
}

\maketitle

\begin{abstract}
	Few-shot learning allows machines to classify novel classes using only a few labeled samples. Recently, few-shot segmentation aiming at semantic segmentation on low sample data has also seen great interest. In this paper, we propose a learnable module for few-shot segmentation, the task-adaptive feature transformer (TAFT). TAFT linearly transforms task-specific high-level features to a set of task-agnostic features well-suited to the segmentation job. Using this task-conditioned feature transformation, the model is shown to effectively utilize the semantic information in novel classes to generate tight segmentation masks. The proposed TAFT module can be easily plugged into existing semantic segmentation algorithms to achieve few-shot segmentation capability with only a few added parameters. We combine TAFT with Deeplab V3+, a well-known segmentation architecture; experiments on the PASCAL-$5^i$ dataset confirm that this combination successfully adds few-shot learning capability to the segmentation algorithm, achieving the state-of-the-art few-shot segmentation performance in some key representative cases.
\end{abstract}

\section{Introduction}

Deep neural networks have made significant advances in computer vision tasks such as image classification \cite{Resnet, Alexnet, googlenet, inception}, object detection \cite{YOLO, FasterRCNN}, and semantic segmentation \cite{DeepLab, FCN, UNet}. However, training a deep neural network requires a large amount of labeled data, which are scarce or expensive in many cases. Few-shot learning algorithms aim to tackle this problem. Advances in few-shot learning allows machines to handle previously unseen classification tasks with only a few labeled samples in some cases \cite{MAML,PN,MN,TapNet}.

Recently, more complicated few-shot learning problems such as few-shot object detection \cite{FSD1, FSD2, FSD3} and few-shot semantic segmentation \cite{OSLSM, coFCN, CANet} have seen much interest. Labels for object detection or semantic segmentation are even harder to obtain, naturally occasioning the formulation of few-shot detection and few-shot segmentation problems. 
In this paper, we tackle the few-shot semantic segmentation problem. The goal of few-shot segmentation is to conduct semantic segmentation on previously unseen classes with only a few examples. Few-shot segmentation tends to be even more challenging than few-shot classification, since segmentation labels contain more information than classification labels.

Specifically, we propose a learnable module, a task-adaptive feature transformer (TAFT), for few-shot segmentation. TAFT can be plugged into existing semantic segmentation algorithms employing the encoder-decoder structure, to add few-shot learning capability.
In such segmentation algorithms, the encoder extracts features from input images, and the decoder generates the segmentation masks using the features. Although multi-scale features can be utilized, it is the high level features that contain the most semantic information. The proposed TAFT method is able to convert the semantic information from novel classes into a form that is more comprehensible to the decoder, by transforming the high-level feature to task-agnostic features that contain the information sufficient for segmentation regardless of the given task.	
Few-shot segmentation can be done effectively by the decoder generating the segmentation mask based on these task-agnostic features.


The transformation matrix, which is the only moving part that changes according to the given task, realizes linear transformation that brings all the class prototypes in the embedded feature space close to another set of task-independent class references in the task-agnostic feature space.
These task-independent reference vectors are meta-learned with the update taking place at the end of every episode processing stage, but are completely decoupled from the current task. In contrast, the prototypes that are driven by the input and naturally undergo significant changes from one task to next. Even during meta-learning, as the input task changes, the reference vectors tend to adjust themselves only a little by little.   
For a given task, the transformation converts the pixels of the high-level feature so that each pixel is brought close to the relatively stationary reference vector of the corresponding class. Task-conditioning is provided by this feature transformation.
The decoder can more easily generate the segmentation mask based on the transformed task-agnostic features.



We evaluate the few-shot learning capability of the proposed TAFT module by combining it with the well-known segmentation algorithm, Deeplab V3+.
On 5-shot testing using the PASCAL-$5^i$ dataset, TAFT plugged into Deeplab V3+ shows the state-of-the-art performance on both the mean Intersection-over-Union (mIoU) and binary Intersection-over-Union (binary IoU) scores. On 1-shot testing, the proposed combination yields the state-of-the-art performance on the binary IoU score, although not on the mIoU score. 

\section{Proposed Task-Adaptive Feature Transformer Method}

\subsection{Task-Adaptive Feature Transformer}

TAFT provides an effective task-conditioning for few-shot segmentation. Like in \cite{PANet,CANet,SG}, TAFT utilizes class prototypes $\mathbf{c}_{k}$, which are the pixel averages of the embedded features from the support images for individual classes. 
Unlike in previous few-shot segmentation methods, however, TAFT also makes use of another set of class references, $\mathbf{r}_{k}$. 
While $\mathbf{c}_{k}$ are apparently driven by and thus depend on the current task's input images, $\mathbf{r}_{k}$ are completely decoupled from the current task.

To understand the role of $\mathbf{r}_{k}$, we must describe the meta-learning process. Assume an episodic training where each new task or episode of a support set and a query set is presented to the model one at a time. As an episodic processing stage ends, encoder $f$, decoder $g$ and this class reference vector set $\mathbf{r}_{k}$ are updated and passed on to the next stage. With a new episode, prototypes $\mathbf{c}_{k}$ are computed using $f$ and the support set. The linear transformation matrix $\mathbf{P}$ is then found 
such that the transformed prototype $\mathbf{Pc}_{k}$ is brought near $\mathbf{r}_{k}$ for all $k$'s. Now as the query set is processed, the corresponding encoder output features get transformed pixel-by-pixel via $\mathbf{P}$. Loss is computed that depends in part on the distances between the pixels of the transformed query features and the references $\mathbf{r}_{k}$. The decoder predicts the segmentation masks of the query images using the transformed query features. Another loss term is then computed based on the predicted segmentation masks. The encoder, decoder and $\mathbf{r}_{k}$ get updated based on these losses. 

We observe that while the prototypes $\mathbf{c}_{k}$ may change significantly from one task to next as inputs vary, the other class reference set $\mathbf{r}_{k}$ changes only by little from one episode to next. As such, $\mathbf{r}_{k}$ offers a stable classification space, while the linear transformation $\mathbf{P}$, which is constructed anew in every episode, provides quick task-conditioning. The decoder makes use of the query features transformed to the space where $\mathbf{r}_{k}$'s reside. In this sense, the stability of  $\mathbf{r}_{k}$ helps the decoder to steadily develop a strategy to generate the segmentation mask.

\begin{figure*}[h]
	\centering
	\includegraphics[width=0.90\textwidth]{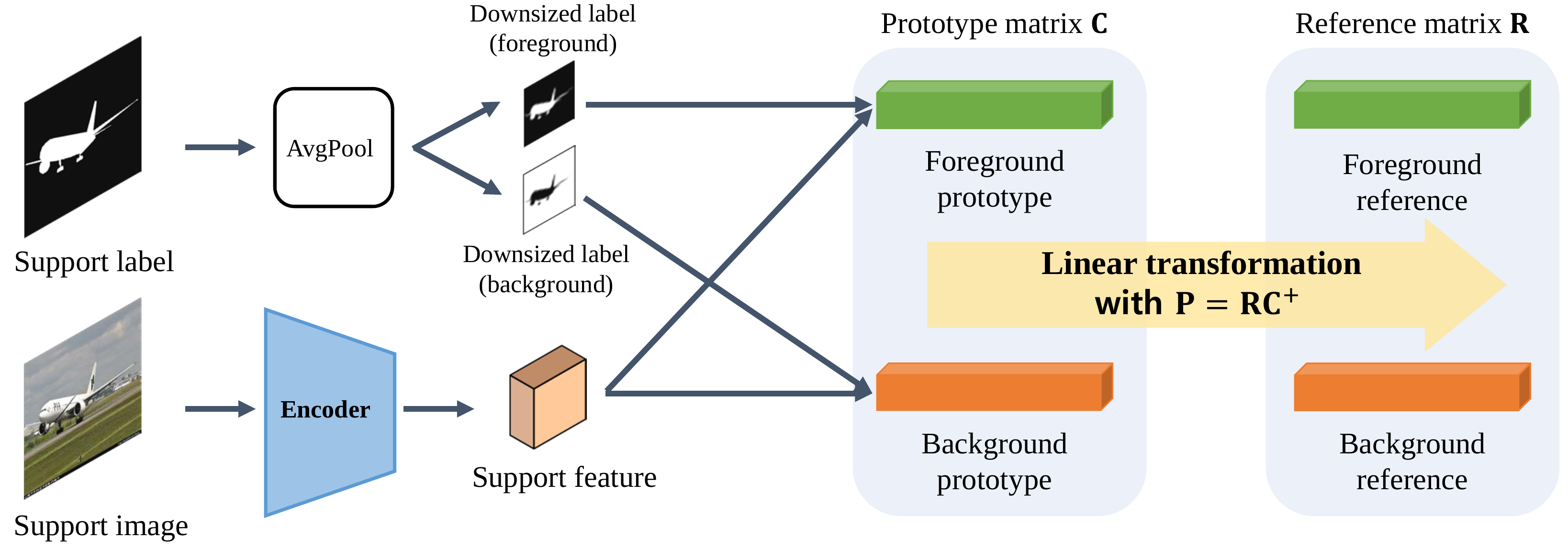}
	\caption{Process of constructing transformation matrix in TAFT}
	
	\label{figure1}
\end{figure*}

Figure \ref{figure1} visualizes the TAFT process that constructs the transformation matrix for 1-way 1-shot segmentation. The reference vectors $\{\mathbf{r}_{fg},\mathbf{r}_{\text{bg}}\}$ for foreground and background having the length matching the number of channels in high-level feature are utilized. 
First, the encoder $f$ extracts the high-level feature $\mathbf{h} = f(\mathbf{x})$ from the support image $\mathbf{x}$. The downsized labels $\tilde{y}_{fg}$ and $\tilde{y}_{\text{bg}}$ having the same size as the high-level feature are generated by average pooling. Although we can use a simple resizing when generating the downsized labels, resizing often leads to binary hard labels with all foreground or all background when the object in the image is too large or too small; this is problematic in prototype computation. On the other hand, average pooling generates soft labels with continuous values between 0 and 1, and we can compute both foreground and background prototypes for all cases.   
Using the downsized soft label $\tilde{y}_{fg}$ and feature $\mathbf{h}$, the foreground prototype $\mathbf{c}_{fg}$ is computed as

\begin{equation}
\mathbf{c}_{\text{fg}} = \frac{\sum_{i}\sum_{j}\mathbf{h}^{(i,j)}\tilde{y}_{\text{fg}}^{(i,j)}}{\sum_{i}\sum_{j}\tilde{y}_{\text{fg}}^{(i,j)}}
\label{eq_1}
\end{equation}

where $\mathbf{h}^{(i,j)}$ and $\tilde{y}_{fg}^{(i,j)}$ denote the pixels in $\mathbf{h}$ and $\tilde{y}_{fg}$, respectively. Likewise, using soft label $\tilde{y}_{\text{bg}}$ and feature $\mathbf{h}$ the background prototype $\mathbf{c}_{\text{bg}}$ is computed as

\begin{equation}
\mathbf{c}_{\text{bg}} = \frac{\sum_{i}\sum_{j}\mathbf{h}^{(i,j)}\tilde{y}_{\text{bg}}^{(i,j)}}{\sum_{i}\sum_{j}\tilde{y}_{\text{bg}}^{(i,j)}}
\label{eq_2}
\end{equation}

where $\tilde{y}_{\text{bg}}^{(i,j)}$ are the pixels in $\tilde{y}_{\text{bg}}$.
For $N$-shot setting with $N$ support images and support labels, the prototypes $\mathbf{c}_{\text{fg}}, \mathbf{c}_{\text{bg}}$ are computed as the means of the sample prototypes computed by equations (\ref{eq_1}) and (\ref{eq_2}). Given the prototypes $\{\mathbf{c}_{\text{fg}},\mathbf{c}_{\text{bg}} \}$ and reference vectors $\{\mathbf{r}_{\text{fg}},\mathbf{r}_{\text{bg}} \}$, we compute the prototype matrix $\mathbf{C}$ as $\left[\frac{\mathbf{c}_{\text{fg}}}{\norm{\mathbf{c}_{\text{fg}}}},\frac{\mathbf{c}_{\text{bg}}}{\norm{\mathbf{c}_{\text{bg}}}}\right]$ and reference matrix $\mathbf{R}$ as $\left[\frac{\mathbf{r}_{\text{fg}}}{\norm{\mathbf{r}_{\text{fg}}}},\frac{\mathbf{r}_{\text{bg}}}{\norm{\mathbf{r}_{\text{bg}}}}\right]$. For general $M$-way segmentation, these matrices will have $M$ columns for foreground classes and the same single column for background. 

We can construct the transformation matrix $\mathbf{P}$ by finding a matrix such that $\mathbf{PC}=\mathbf{R}$. In general, $\mathbf{C}$ is not a squre matrix and does not have an inverse. One way to find a reasonable $\mathbf{P}$ is to compute 
$\mathbf{P} = \mathbf{R}\mathbf{C}^+$, where $\mathbf{C}^+$ is the pseudo-inverse of $\mathbf{C}$ computed as $\{\mathbf{C}^{T}\mathbf{C}\}^{-1}\mathbf{C}^{T}$. This gives the least square fit between $\mathbf{PC}$ and $\mathbf{R}$. 
Matrix $\mathbf{P}$ depends on $\mathbf{C}$, which is driven by the encoder input representing the current task. Thus, it can be said that task-conditioning is achieved via application of the linear transformation using $\mathbf{P}$.
Given the high-level feature from a query image, TAFT transforms it to the task-agnostic feature pixel-by-pixel using $\mathbf{P}$. 
The pixel-wise feature transformation can be easily done using a $1 \times 1$ convolution layer with weight $\mathbf{P}$. 
With this task-adaptive feature transformation, the pixels of the task-agnostic feature get to settle close to the corresponding reference vectors, and the decoder can easily distinguish the foreground and background pixels in the task-agnostic feature. Note that the reference vectors are the only learning part in TAFT. The transformation matrix is not trained but computed for each task.

\subsection{Combination with Segmentation Algorithm}

The TAFT module demonstrates few-shot segmentation capability by being combined with semantic segmentation methods. In Algorithm \ref{alg}, we display the procedure to process a given episode for the TAFT module combined with a segmentation algorithm during meta-training (which is the same as the inference process except for the loss computation and parameter update part). The prototypes are computed using the downsized labels and encoder features from the support set (line 2 to 10). Then the transformation matrix $\mathbf{P}$ is constructed using the prototypes and reference vectors (line 11 to 12). 		
For training the reference vectors, we utilize an auxiliary loss $L_R$. Using the reference vectors and task-agnostic features, we generate the downsized predictions and compare them with downsized labels (line 16). In prediction, we use the pixel-wise inner product with the reference vectors and normalize the prediction scores for each pixel using the softmax activation function. Since the downsized labels contain continuous values in $[0,1]$, the mean-squared-error (MSE) loss for regression is utilized as the auxiliary loss and computed over all pixels of downsized predictions. Using $L_R$, the reference vectors are trained so that the pixels in task-agnostic features can be easily distinguished. 
Using the task-agnostic features, the decoder predicts the segmentation masks (line 17). The segmentation loss $L_S$ is computed between the labels and the predicted segmentation masks (line 18). The segmentation loss is a cross-entropy loss computed over all pixels in images, which is generally used for semantic segmentation. Through meta-learning using $L_S$, the decoder learns to generate segmentation masks using the task-agnostic features where the each pixel is close to the corresponding reference vector. 

\begin{algorithm*}[h]
	\caption{Procedure for TAFT combined with the encoder $f$ and decoder $g$ of the segmentation algorithm to process an episode during meta-training. The episode is composed of $N_{s}$ support images/labels and $N_{q}$ query images/labels. Label $\mathbf{y}_k$ is composed of foreground label $y_{k,fg}$ and  background label $y_{k,bg}$. $\mathbf{r}_{\text{fg}}$ and  $\mathbf{r}_{\text{bg}}$ are the reference vectors for foreground and background, respectively, and $y^{(i,j)}$ and  $\mathbf{h}^{(i,j)}$ denote the $(i,j)$th pixel of $y$ and $\mathbf{h}$, respectively.}
	\label{alg}
	\textbf{Input}: Encoder $f$, decoder $g$, reference vectors $[\mathbf{r}_{\text{fg}},\mathbf{r}_{\text{bg}}]$,  \\$S = \{(\mathbf{x}_{1}, \mathbf{y}_{1}),...,(\mathbf{x}_{N_s}, \mathbf{y}_{N_s})\}$, $Q = \{(\bar{\mathbf{x}}_{1},\bar{\mathbf{y}}_{1}),...,(\bar{\mathbf{x}}_{N_q}, \bar{\mathbf{y}}_{N_q})\}$ 
	
	\begin{algorithmic}[1]
		\STATE $L_s\leftarrow0,\;\;\;\; L_r\leftarrow0 $
		\FOR{$n$ in $ \left \{1 , ... , N_s \right \}$}
		\FOR{$k$ in $ \left \{\text{fg,bg} \right \}$}
		\STATE $ \tilde{y}_{n,k} \leftarrow \text{AvgPool}(y_{n,k})$\ \ \ \ \ \ \ \ \ \ \ \quad $\triangleright$ \ $ y_{n,k} \in \mathbb{R}^{(H \times W)},\;\; \tilde{y}_{n,k} \in \mathbb{R}^{(H_s \times W_s)} $
		
		\STATE $\mathbf{c}_{n,k}\leftarrow \sum_{i}\sum_{j}f(\mathbf{x}_n)^{(i,j)}\tilde{y}_{n,k}^{(i,j)}/\sum_{i}\sum_{j}\tilde{y}_{n,k}^{(i,j)}$
		\ENDFOR
		\ENDFOR
		\FOR{$k$ in $ \left \{\text{fg,bg} \right \}$}
		\STATE $\mathbf{c}_{k} \leftarrow \sum_n{\mathbf{c}_{n,k}}/N_s $
		\ENDFOR
		\STATE $\mathbf{C} \leftarrow [\mathbf{c}_{\text{fg}}/\norm{\mathbf{c}_{\text{fg}}},\mathbf{c}_{\text{bg}}/\norm{\mathbf{c}_{\text{bg}}}], \;\;\;\; \mathbf{R} \leftarrow [\mathbf{r}_{\text{fg}}/\norm{\mathbf{r}_{\text{fg}}},\mathbf{r}_{\text{bg}}/\norm{\mathbf{r}_{\text{bg}}}] $
		
		\STATE $\mathbf{P} \leftarrow \mathbf{R}\{\mathbf{C}^{T}\mathbf{C}\}^{-1}\mathbf{C}^{T}$
		
		\FOR{$n$ in $ \left \{1 , ... , N_q \right \}$}
		\STATE $\mathbf{h}_a^{(i,j)} \leftarrow \mathbf{P}f(\mathbf{\bar{x}}_n)^{(i,j)}$ \ \ \ \ \ \ \ \ \ \ \ \ \ \ \ \ \quad $\triangleright$ $\mathbf{h}_a$ denotes the task-agnostic feature
		\STATE $ \tilde{y}_{n,fg} \leftarrow \text{AvgPool}(\bar{y}_{n,fg}), \;\;\;\; \tilde{y}_{n,bg} \leftarrow \text{AvgPool}(\bar{y}_{n,bg}) $
		
		\STATE $L_{R} \leftarrow L_{R} + \displaystyle\frac{1}{2\times H_s\times W_s} \displaystyle\sum_{k\in\{\text{fg,bg}\}}\displaystyle\sum_{i}\sum_{j}\text{MSE}\Big(\tilde{y}_{n,k}^{(i,j)}, \frac{\exp\big(\mathbf{r}_{k}\mathbf{h}_a^{(i,j)}\big)}{\exp\big(\mathbf{r}_{\text{fg}}\mathbf{h}_a^{(i,j)}\big)+\exp\big(\mathbf{r}_{\text{bg}}\mathbf{h}_a^{(i,j)}\big)}\Big)$
		\STATE $\hat{y}_n \leftarrow g(\mathbf{h}_a)$ 
		\STATE $L_{S} \leftarrow L_{S} + \displaystyle\frac{1}{ H\times W} \sum_{i}\sum_{j}\text{CrossEntropy}(y_{n,fg}^{(i,j)},\hat{y}_{n}^{(i,j)}) $
		\ENDFOR
		\STATE Update $f$ minimizing $L_r+L_s$; update $g$ minimizing $L_s$; update $[\mathbf{r}_{\text{fg}},\mathbf{r}_{\text{bg}}]$ minimizing $L_r$.
	\end{algorithmic}
	\textbf{Output}: Updated encoder $f$, decoder $g$, reference vectors  $[\mathbf{r}_{\text{fg}},\mathbf{r}_{\text{bg}}]$
\end{algorithm*}

\begin{figure*}[h]
	\centering
	\includegraphics[width=0.90\textwidth]{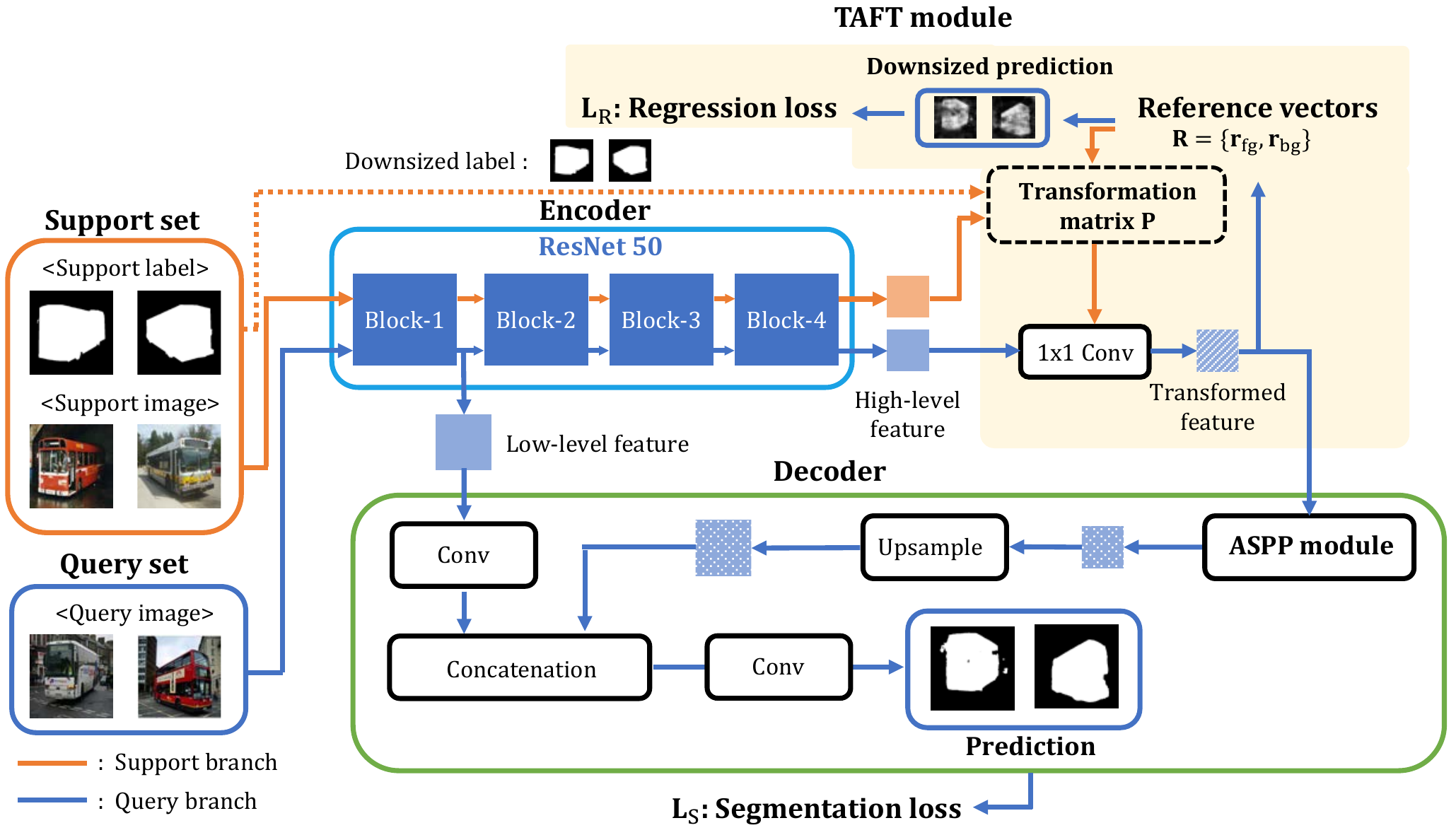}
	\caption{Architecture of Deeplab V3+ combined with TAFT}
	
	\label{figure3}
\end{figure*}

In this paper, we test the TAFT module in conjunction with Deeplab V3+. Deeplab V3+ of \cite{FCN} is a semantic segmentation algorithm showing the best performance in the VOC 2012 segmentation dataset.  
Deeplab V3+ consists of the encoder, decoder, and the Atrous Spatial Pyramid Pooling (ASPP) module. For segmentation, the high-level feature is extracted from the image by the encoder, and the low-level feature is extracted together from the middle layer of the encoder. The high-level feature is then processed by the ASPP module to capture the multi-scale information, and the decoder network generates the segmentation mask using the ASPP output feature and the low-level feature. In Deeplab V3+, the high-level feature contains the semantic information while the low-level feature contains the shape information. Both features with different information contents are used together to predict the segmentation mask.

Figure \ref{figure3} illustrates how the TAFT module is plugged in Deeplab V3+. TAFT operates on the high-level feature from the encoder and generates the task-agnostic feature. The task-agnostic feature is then processed by the ASPP module, which is included as a part of the decoder in this figure. The decoder generates the segmentation mask using the low-level feature and the ASPP-processed task-agnostic feature together.
Note that TAFT does not process the low-level feature, since the feature contains the shape information which can be considered general.

\begin{figure*}[h]
	\centering
	\includegraphics[width=0.9\textwidth]{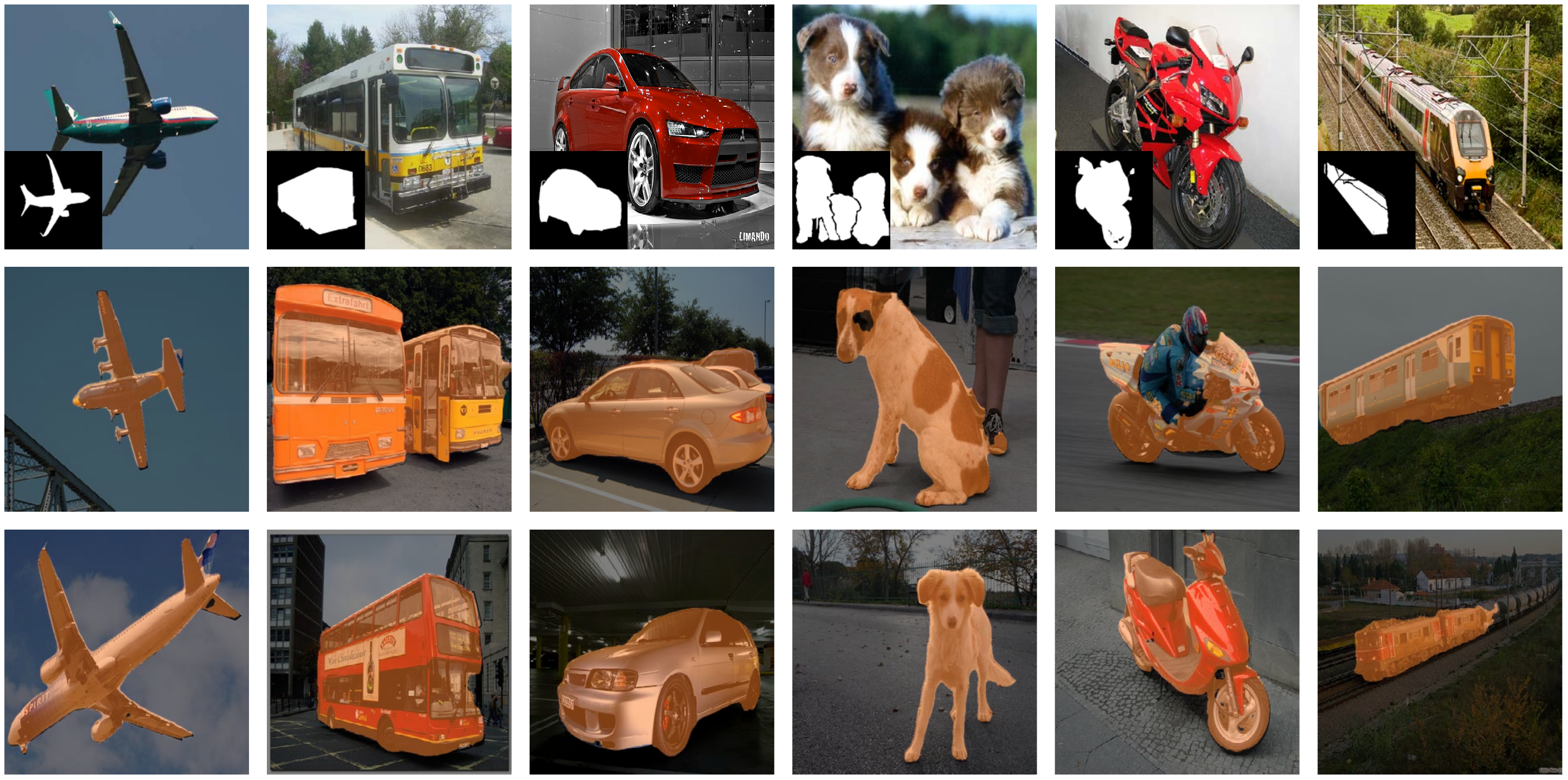}
	\caption{Qualitative results of TAFT + Deeplab V3+ for 1-shot segmentation}
	\label{figure4}
\end{figure*}

Figure \ref{figure4} displays some qualitative results of TAFT combined with Deeplab V3+. We visualize the 1-shot segmentation results of aeroplane, bus, car, dog, motorbike and train classes. The images in the first row are support images, with the support labels shown together at the bottom left of the support images. The images in the second and third rows show the query images with the prediction results. We can see that TAFT combined with Deeplab V3+ successfully segments the objects from the query images using only a single support sample in each class. More qualitative results can be found in Supplementary Material. 

\section{Related Work}

\subsection{Few-Shot Learning}

Few-shot learning aims to develop a general classifier that can quickly adapt to the unseen task with only a few labeled samples, so that it can classify the samples from new classes without much new training. In carrying out few-shot classification, two main strategies have been developed.

One popular way is metric-based few-shot learning. 
The goal of metric-based few-shot learning is to learn a mapping to the metric space where samples from the same category approach together while those from different categories are kept far apart.
For example, Matching Networks of \cite{MN} learn two separate embedding spaces and dynamically extracts semantic features from the support set and query set using bi-directional LSTM. Classification in Matching Networks is done based on cosine correlation between features.
Prototypical Networks of \cite{PN}, on the other hand, are based on the assumption that features of the samples that belong to the same category will cluster into a specific point or a prototype. 
Prototype is defined as the average of support sample features, and classifier makes prediction using the Euclidean distance between query samples and the prototype of each category. 
Based on a Prototypical Networks, TADAM of \cite{TADAM} proposes a task-conditioning algorithm based on element-wise feature scaling and shifting. The task-conditioned feature extractor generates a task-adaptive metric space where actual classification is carried out.
TapNet of \cite{TapNet} suggests another type of task-conditioning by employing task-adaptive projection. 
The projection space is constructed by nulling the error between prototypes and learnable reference vectors for each episode. Actual classification takes place in this task-conditioned space, which amplifies discriminability among features from different classes. 

Another prominent direction is the optimization-based meta-learner. 
This approach aims to train a meta-learner which in turn trains an actual learner. The meta-learner supports learning of the actual learner so that the learner can adapt well to a new task using only a small number of updates using a few samples. The meta-learner LSTM of \cite{Ravi} trains an LSTM as a meta-learner to optimize another learner. The parameters of another learner are trained based on the memory update rule of LSTM, effectively adapting to a new task with only a few labeled samples. 
In a concurrent work, MAML of \cite{MAML} trains a model initialization which can be easily fine-tuned using only a few gradient steps. To achieve this, the initial model is learned by aggregating gradients from multiple tasks to increase the sensitivity of the model parameters to a new task. The idea of MAML influenced many variants such as \cite{TAML,REPTILE, LEO}.


Among the few-shot learning algorithms, TAFT is most closely related to TapNet of \cite{TapNet} in that
stand-alone meta-learned reference vectors are employed and linear transformation is used. Despite the similarity, there are some key differences in both philosophical viewpoints as well as design methodologies. In TAFT, the reference vectors are viewed as a more stable form of prototypes residing in a task-agnostic space, which serves as the destination for linear transformation. TAFT brings the embedded features all the way to this task-agnostic space, where inference and reference updates take place. This tends to make the references in TAFT more stable than those in TapNet, which helps reliable generation of segmentation masks in the decoder and subsequent training of decoder modules. TapNet lacks this interpretation as the projection space, where its inference and subsequent reference updates are done, is some intermediate distance away from both the prototypes and the classification references. In other words, TapNet's reference vectors themselves are also the subject of transformation. As for specific design methodology, TapNet's linear transformation attempts to align in-class pairs of vectors while at the same time distancing out-class vectors as much as possible, to maximize classification accuracy. TapNet employs linear nulling of errors for this purpose. In comparison, for TAFT, the least square fit criterion is  more appropriate for linear transformation, as it is well-suited for segmentation.

\subsection{Few-Shot Segmentation}

Few-shot segmentation attempts semantic segmentation on novel classes using only a few examples. 
OSLSM of \cite{OSLSM} is a first algorithm adopting the few-shot learning strategy to semantic segmentation. OSLSM utilizes a two-branch structure involving a conditioning branch and a segmentation branch. The conditioning branch processes the support set and generates the scaling and shifting parameters, and the segmentation branch is task-conditioned by element-wise scaling and shifting with these parameters.
Co-FCN of \cite{coFCN} also utilizes a two-branch architecture. The conditioning branch generates globally pooled prediction and the segmentation branch fuses it with query features in predicting the segmentation mask. Conditioning with the globally pooled prediction shows robust performance for sparsely annotated samples. 

Recent methods utilize the prototype idea of \cite{PN}, to use the information of support set efficiently. 
SG-One of \cite{SG} utilizes the masked-average pooling to compute the prototypes from the feature pixels of support samples. The similarity map is generated from the cosine-similarity results between the prototypes and feature pixels of query sample, and utilized to guide segmentation. 
CANet of \cite{CANet} also utilizes the prototype concept, generating a prototype for a given class with masked-average pooling. The prototype is concatenated with every feature pixel of query for dense comparison. An optimization module that iteratively updates segmentation mask prediction is also proposed. 
PANet of \cite{PANet} again utilizes masked-average pooling to construct prototypes. The segmentation is done by classifying each pixel to the class of the nearest prototype. For more effective meta-training, a novel regularization method, prototype alignment, is proposed. Prototype alignment is done by reverse direction prediction that predicts the segmentation mask of support samples using the prediction results of query images as the labels of query samples during meta-training. 

Instead of using prototypes, PGNet of \cite{PGNet} models support and query features as graphs where nodes correspond to regions of the features, and combination between the support graphs and query graphs is modeled as a big bipartite graph. The graph attention mechanism generates the connection weight between the nodes. Moreover, pyramid graph reasoning that relies on multiple graphs with different sizes of regions corresponding to the node is utilized to capture the multi-level relationship.

Our TAFT is similar to the works of \cite{PANet,CANet,SG} in the sense that the prototypes are utilized to represent the pixels. The prior works directly utilize the prototypes that vary widely from task to task in comparing the pixels of the feature with the prototypes. 
However, the proposed TAFT method does not rely on comparison between the prototypes and the pixels of feature. Instead, TAFT utilizes feature transformation converting task-specific features into fairly stable, task-agnostic features. 
While the prototype-based approaches require sample-specific comparison between the prototypes and pixels of the feature for each image sample, TAFT does not require any sample-specific process; it generates the transformation matrix $\mathbf{P}$ and use it for every query image. This enables an efficient parallel computation to predict the segmentation masks of different samples at the same time. 
Moreover, unlike the methods of \cite{CANet,SG}, TAFT does not require additional learnable parts except the reference vectors, which require only a small number of parameters. These characteristics make TAFT easy to combine with existing segmentation algorithms.

\section{Experiment Results}

\subsection{Dataset and Evaluation Metric}
\textbf{PASCAL-}$\mathbf{5^i}$ is used for our experiments. PASCAL-$5^i$ is a dataset based on PASCAL VOC 2012, proposed by \cite{OSLSM} for few-shot segmentation. The 20 classes of VOC 2012 is divided into 4 splits, and each split contains 5 disjoint classes. In our experiments, the 5 classes in one of 4 splits are selected as the test classes, and remaining classes are used as the training classes. The training samples of training classes and the test samples of test classes are used for training and testing, respectively. For example, in the experiment with split 0 of PASCAL-$5^i$, training is done with the training samples of 15 classes in splits 1,2, and 3, while evaluation is done with the test samples of 5 classes in split 0. The details of class split can be found in \cite{OSLSM}. Data augmentation is not applied.

We use the mean Intersection-over-Union(mIoU) score as the evaluation metric. As done in prior works, we compute the foreground IoU for each class, and use the averaged per-class foreground IoU as mIoU. We also report the evaluation result with another metric, the binary IoU score suggested in \cite{coFCN}. The binary IoU score is computed as the average of foreground IoU and background IoU computed over all test images. 

\subsection{Experimental Settings}

For training TAFT+Deeplab V3+, we utilize the Adam optimizer of \cite{Adam} with a learning rate of $10^{-4}$. We initialize the encoder with the ImageNet pretrained ResNet-50, and we apply a 10 times smaller learning rate to the encoder as done in Deeplab V3+. 
We modified the strides in the ResNet-50 encoder so that the low-level feature and high-level feature are downsized by a factor of 4 and 16, respectively. Since PASCAL-$5^i$ includes many small objects, we use the optimized atrous rates of $[1,4,7,11]$ in the  ASPP module. 

For the 1-shot cases, $3\times10^4$ episodes are used for meta-training and the learning rate is decayed by a factor of 10 after training $2 \times 10^{4}$ episodes. The weight decay of \cite{AdamW} with the optimized decay rate is applied for regularization. Training is done using episodes with 12 queries.
For the 5-shot experiments, the model is meta-trained with $3\times10^4$ episodes and a learning rate decay by a factor of 10 is applied after training $2 \times 10^{4}$ episodes. The $l2$ weight decay of \cite{AdamW} with the optimized rate is applied for training the encoder. The episodes with 10 query samples are used for training.
More detailed hyperparameter settings can be found in Supplementary Material.    

In both training and evaluating in the 1-shot and 5-shot cases, we use the images and labels resized to $512 \times 512$, considering the crop size used in Deeplab V3+. In evaluation, the episodes with 5 queries are used. The model is evaluated by 6,000 randomly generated episodes per class. Since each split contains 5 classes, 30,000 episodes are used for evaluation in total. As done in prior works of \cite{PGNet,CANet}, the multi-scale input test with scale factors $[0.7, 1, 1.3]$ is used. In evaluation, the support and query samples are scaled with a specific ratio, and the corresponding predictions are scaled again into the original ratio. The final prediction is done by averaging 3 predictions from different scales. 

\subsection{Experimental Results}

We compare the proposed TAFT + Deeplab V3+ method with prior approaches. In Tables \ref{table1}, for each split and the mean, we display the mIoU scores for both 1-shot and 5-shot segmentations on left and right, respectively. 
For 1-shot, TAFT + Deeplab V3+ achieves somewhat lower mIoU scores compared to two prior methods: CANet and PGNet. 
In 5-shot segmentation, on the other hand, TAFT + Deeplab V3+ exhibits significantly higher mIoU scores than all of the prior methods. While the difference between the scores of two previous best methods is rather small at 1.4 in the mean, the proposed method improves the previous best by 5.0, a large margin. 

\begin{table*}[h]
	\centering
	\begin{tabular}{c|cc|cc|cc|cc|cc}
		\toprule  
		\textbf{Models}    & \multicolumn{2}{c|}{split-0} & \multicolumn{2}{c|}{split-1}	& \multicolumn{2}{c|}{split-2} &\multicolumn{2}{c|}{split-3} &\multicolumn{2}{c}{mean} \\
		\midrule
		\textbf{OSLSM} \cite{OSLSM}   & 33.6 &35.9	&55.3  &58.1& 40.9 &42.7  & 33.5 &39.1 & 40.8 &43.9  \\
		\textbf{co-FCN} \cite{coFCN}  & 36.7 &37.5	&50.6 &50.0 & 44.9 &44.1  &32.4 &33.9 & 41.1  &41.4 \\
		\textbf{SG-One} \cite{SG}  & 40.2 &41.9	&58.4 &58.6 & 48.4 &48.6  & 38.4 &39.4& 46.3 &47.1  \\
		\textbf{CANet} \cite{CANet} & 52.5 &55.5	&65.9 &67.8 &\textbf{51.3}  &51.9& \textbf{51.9} &53.2 & 55.4 &57.1  \\
		\textbf{PANet} \cite{PANet} & 42.3 &51.8	&58.0 &64.6 & 51.1  &59.8& 41.2 &46.5 & 48.1 &55.7  \\
		\textbf{PGNet} \cite{PGNet}  & \textbf{56.0} &57.9 &\textbf{66.9}  &68.7& 50.6  &52.9& 50.4 &54.6 & \textbf{56.0} &58.5  \\
		\midrule
		\textbf{TAFT + Deeplab V3+}  &50.2 &\textbf{59.4}	&57.9 &\textbf{69.5}&50.9 &\textbf{64.5}  & 49.4 &\textbf{60.7} &52.1 &\textbf{63.5}  \\
		\bottomrule
	\end{tabular}
	\caption{1-shot (left) and 5-shot (right) mean Intersection-over-Union (mIoU) scores for PASCAL-$5^i$}
	\label{table1}
\end{table*}


Table \ref{table3} shows the binary IoU scores (only one past work reported results for all splits). For this criterion, TAFT + Deeplab V3+ shows the best binary IoU mean score for both 1-shot and 5-shot. For 1-shot, the gain over the previous best is small, but for 5-shot, our method again shows a large improvement relative to the current state-of-the-art methods. Two previous best methods differ only by 0.2 in the mean score, whereas the proposed method gives a 6.1 improvement in score over the previous best. The higher binary IoU score means that the proposed method achieves higher background IoU than prior works. We conjecture that the usage of background pixel information results in the higher background IoU and binary IoU scores, as in PANet of \cite{PANet}. While most prior works only utilize the foreground prototype or foreground pixel information, 
TAFT utilizes the prototypes and reference vectors for both foreground and background. 


\begin{table*}[h]
	\centering
	\begin{tabular}{c|cc|cc|cc|cc|cc}
		\toprule  
		\textbf{Models}    & \multicolumn{2}{c|}{split-0} & \multicolumn{2}{c|}{split-1}	& \multicolumn{2}{c|}{split-2} &\multicolumn{2}{c|}{split-3} &\multicolumn{2}{c}{mean} \\
		\midrule
		\textbf{OSLSM} \cite{OSLSM}   & - &-	&- &-& - &-  & - &- & 61.3 &61.5 \\
		\textbf{co-FCN} \cite{coFCN}  & - &-	&- &-& - &-  & - &- &60.1 &60.2  \ \\
		\textbf{SG-One} \cite{SG}  & - &-	&- &-& - &-  & - &- & 63.1 &65.9  \ \\
		\textbf{CANet} \cite{CANet}  &\textbf{71.0} &\textbf{74.2}&\textbf{76.7} &80.3& 54.0 &57.0 & 67.2 &66.8 & 66.2 &69.6 \  \\
		\textbf{PANet} \cite{PANet} & - &-	&- &-& - &-  & - &- & 66.5 &70.7  \  \\
		\textbf{PGNet} \cite{PGNet}  & - &-	&- &-& - &-  & - &- & 69.9 &70.5  \ \\
		\midrule
		\textbf{TAFT + Deeplab V3+}   & 69.8 &74.0	&73.4&\textbf{81.0}&\textbf{ 68.9}&\textbf{76.0}  & \textbf{67.9}&\textbf{76.0} & \textbf{70.0} &\textbf{76.8}  \  \\
		\bottomrule
	\end{tabular}
	\caption{1-shot (left) and 5-shot (right) binary Intersection-over-Union (IoU) scores for PASCAL-$5^i$}
	\label{table3}
\end{table*}


\section{Conclusion}
In this paper, we propose a task-adaptive feature transformer module for few-shot segmentation. TAFT transforms the high-level feature containing the semantic information into a task-agnostic feature. 
The feature transformation converts the semantic information in the high-level feature to a form more suitable for the decoder to generate the segmentation mask. The reference vectors that are meta-trained and known to the decoder module are utilized as the target for feature transformation. The pixels of the feature are then brought closer to the reference vector of the corresponding class by this feature transformation.
The proposed TAFT module can be easily combined with existing semantic segmentation algorithms such as Deeplab V3+, to give few-shot learning capability to them.
Extensive experiments on PASCAL-$5^i$ show that TAFT plugged into Deeplab V3+ achieves the state-of-the-art few-shot segmentation performance in some key test cases.

{\small
	\bibliographystyle{ieee_fullname}
	\bibliography{egbib}
}
\clearpage


\section*{Supplementary material}

# Task-Adaptive Feature Transformer for Few-Shot Segmentation: Supplementary Material

Jun Seo
KAIST
tjwns0630@kaist.ac.kr

Young-Hyun Park
KAIST
dnffkf369@kaist.ac.kr

Sung-Whan Yoon
UNIST
shyoon8@unist.ac.kr

Jaekyun Moon
KAIST
jmoon@kaist.edu

## 1. Hyperparameter Settings

In Table 1, we display the hyperparameter settings for TAFT + Deeplab V3+. We utilize the learning rate of $10^{-4}$ for training the reference vectors, ASPP module, and decoder module for both 1-shot and 5-shot cases. For training of the encoder, a learning rate of $10^{-5}$, which is one tenth of the learning rate for the other modules, is applied. For both cases, we reduce the learning rate by $1/10$ after training 20,000 episodes. During meta-training, the episodes with $N_q$ queries are used as the training episodes. For regularization, we apply different weight decay rates for the encoder and the other modules. $\gamma_1$ is the weight decay rate for the reference vectors, the ASPP module, and the decoder module, and $\gamma_2$ is the decay rate for the encoder. For the segmentation loss, we compute the ratio between the number of foreground and background pixels over whole training sets, and apply that ratio to the class weight of cross entropy loss. We use the scale weights of 1 and 4 for background class and foreground class, respectively.

## 2. Additional Experimental Results

### 2.1. Combination with FCN

In the main paper, we argue that TAFT can be combined with the existing semantic segmentation algorithm achieving few-shot segmentation capability. As an example, we showed that TAFT combined with Deeplab V3+ achieved state-of-the-art few-shot segmentation performance. To further demonstrate the extendibility and versatility of TAFT, we plug TAFT into another well-known segmentation algorithm, Fully Convolutional Network (FCN). FCN utilizes the convolutional neural network (CNN) architectures designed for classification by convolutionalization. The convolutionalization process transforms the fully connected layers as $1 \times 1$ convolution layers. In this way, the network can preserve the location information while predicting the class of each pixel. After prediction, FCN upsamples the scores into the original image size. In FCN, the convolution layers or the feature extractor of classification network is considered as encoder, and the convolutionalized layers and upsampling layers are considered as decoder. Since FCN can utilize different classification networks, there can be many different variations. Among them, we utilize the FCN architecture with the ResNet-50 encoder provided by the pytorch framework in our experiment. Figure 1 illustrates this version FCN combined with the TAFT module. The feature transformation by TAFT is applied to the high-level feature from the ResNet-50 encoder. The downsized prediction is generated with the task-agnostic feature and the reference vectors, and the regression loss is computed with downsized label and prediction. The convolutional decoder generates the segmentation mask with the task-specific high-level feature. The segmentation loss is then computed using the original label and the generated segmentation mask. For training TAFT+FCN, we utilize the same hyperparameters as those of 5-shot TAFT+Deeplab V3+ except the number of queries in training episodes. We employ $N_q = 4$ for training the 5-shot TAFT+FCN method. In Tables 2 and 3, we display the 5-shot mean Intersection-over-Union (mIoU) scores and binary Intersection-over-Union (IoU) of TAFT+FCN in comparison with those of other methods. We can see that TAFT + FCN achieves the best mIoU and binary IoU scores except TAFT + Deeplab V3+. TAFT + FCN outperforms the previous best by 2.6 in mIoU and 4.2 in binary IoU.

### 2.2. Ablation Study on TAFT + Deeplab V3+

In Tables 4 and 5, we evaluate the effect of multi-scale input test in TAFT + Deeplab V3+. For both cases, we observe the performance enhancement due to the multi-scale input test. For 1-shot, the average mIoU and binary IoU scores increase by 0.8 and 0.9, respectively. For 5-shot, on the other hand, the average mIoU and binary IoU scores increase by 1.5 and 1.3. Note that for 5-shot, TAFT + Deeplab V3+ outperforms prior works without multi-scale input test.

In the experiments on TAFT + Deeplab V3+ in the main paper, we do not apply the feature transformation to the low-level feature, since it contains shape information which can

| Model | learning rate | lr decay step | $N_q$ | $\gamma_1$ | $\gamma_2$ |
|---|---|---|---|---|---|
| **TAFT + Deeplab V3+** 1-shot | 1e-4 | 20000 | 12 | 0 | 1e-3 |
| **TAFT + Deeplab V3+** 5-shot | 1e-4 | 20000 | 10 | 0 | 3e-4 |

Table 1: Hyperparameter Settings for PASCAL-$5^i$ experiments

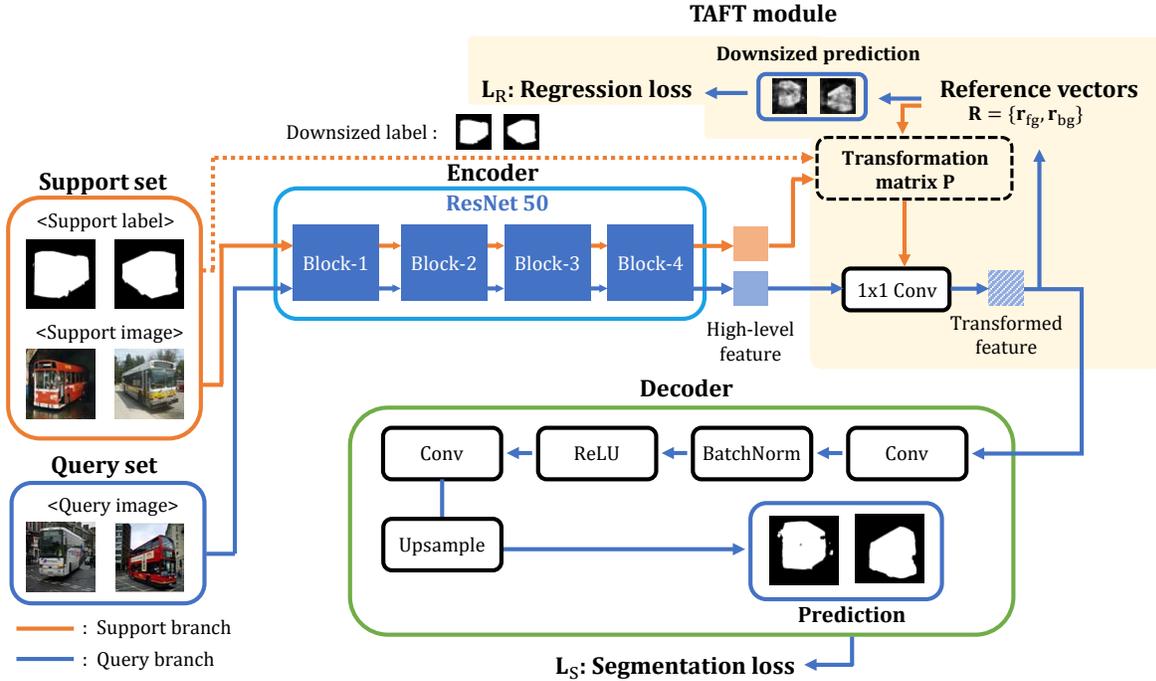

Figure 1: Architecture of TAFT+FCN

be considered general. In Tables 6 and 7, we explore the effect of applying feature transformation to the low-level feature of Deeplab V3+. We apply the low-level feature transformation by adding another reference vector set for low-level feature, and training it using additional auxiliary loss. The transformation matrix is constructed using the additional reference vectors and the prototypes computed from low-level support features. The results in Tables 6 and 7 show that applying feature transformation at low-level feature does not bring any gain in performance; it actually degrades mIoU by 1.1 and binary IoU by 0.5.

### 2.3. Comparison Between Prototype and Reference

In Section 2.1 of the main paper, we stated that the prototypes $c_k$ change significantly while the reference vectors $r_k$ changes only by little. To confirm this, we collect the prototypes and the reference vectors of foreground and background classes from the first episode to the 10,000th episode during meta-training. Then we compute the Euclidean distances $\left\|c_k^{(t)} - c_k^{(t-1)}\right\|$ and $\left\|r_k^{(t)} - r_k^{(t-1)}\right\|$ for $k \in \{fg, bg\}$ and $t \in [2, 10000]$ to measure the changes. In Figure 2, we visualize the percentage changes of the prototypes and reference vectors during meta-training. We can see that the prototypes undergo significant changes while the relative changes in the reference vectors are nearly invisible. Although the class in each episode is randomly sampled and the class may or may not change from one episode to next, the changes in prototype are always much larger than the changes in reference vectors, with or without class changes.

### 2.4. Additional Qualitative Results

In Figures 3 and 4, we display additional qualitative results of TAFT + Deeplab V3+. The 1-shot and 5-shot segmentation results of cat, cow, horse, person, sheep, tv/monitor classes are visualized. As done in the main paper, we show the support set images with the support labels together at the bottom left, and the query images with the prediction results. From these results, we can observe that the quality of segmentation is improved from 1-shot to 5-shot.

2

| Models | split-0 | split-1 | split-2 | split-3 | mean |
|---|---|---|---|---|---|
| OSLSM | 35.9 | 58.1 | 42.7 | 39.1 | 43.9 |
| co-FCN | 37.5 | 50.0 | 44.1 | 33.9 | 41.4 |
| SG-One | 41.9 | 58.6 | 48.6 | 39.4 | 47.1 |
| CANet | 55.5 | 67.8 | 51.9 | 53.2 | 57.1 |
| PANet | 51.8 | 64.6 | 59.8 | 46.5 | 55.7 |
| PGNet | 57.9 | 68.7 | 52.9 | 54.6 | 58.5 |
| TAFT + FCN | 55.4 | 66.9 | 63.2 | 58.7 | 61.1 |
| **TAFT + Deeplab V3+** | **59.4** | **69.5** | **64.5** | **60.7** | **63.5** |

Table 2: 5-shot mean Intersection-over-Union (mIoU) scores for PASCAL-$5^i$

| Models | split-0 | split-1 | split-2 | split-3 | mean |
|---|---|---|---|---|---|
| OSLSM | - | - | - | - | 61.5 |
| co-FCN | - | - | - | - | 60.2 |
| SG-One | - | - | - | - | 65.9 |
| CANet | **74.2** | 80.3 | 57.0 | 66.8 | 69.6 |
| PANet | - | - | - | - | 70.7 |
| PGNet | - | - | - | - | 70.5 |
| TAFT + FCN | 72.1 | 78.4 | 74.7 | 74.3 | 74.9 |
| **TAFT + Deeplab V3+** | 74.0 | **81.0** | **76.0** | **76.0** | **76.8** |

Table 3: 5-shot binary Intersection-over-Union (IoU) scores for PASCAL-$5^i$

## 3. Complexity Comparison with Prior Works

In Section 3.2 of the main paper, we mention that the proposed TAFT module requires only a small amount of learnable parameters. For 1-way segmentation, the number of additional parameters in TAFT is only 4,096. On the other hand, most of the prior methods require considerably more parameters in their key components. Although prior works utilize different base architectures for segmentation, we compare only the number of parameters from their key components. The OSLSM and co-FCN methods utilize the VGG-16 network as the conditioning branch. The number of parameters in convolution layers of VGG-16 is about 15 million, which is a fairly large number. For the SG-One method, the guidance branch consists of 3 convolution layers and the number of parameters in guidance branch is about 1.8 million. For CANet, the dense comparison requires about 1.2 million parameters. The iterative optimization module in CANet also requires some additional parameters. In PGNet, the graph attention unit for comparing support graph and query graph in pyramid graph reasoning requires a large number of additional parameters.

| Models | split-0 | split-1 | split-2 | split-3 | mean |
|---|---|---|---|---|---|
| **TAFT + Deeplab V3+** (1-shot, w/o multi-scale) | 49.0 | 56.9 | 50.9 | 48.4 | 51.3 |
| **TAFT + Deeplab V3+** (1-shot, w/ multi-scale) | 50.2 | 57.9 | 50.9 | 49.4 | 52.1 |
| **TAFT + Deeplab V3+** (5-shot, w/o multi-scale) | 57.5 | 68.2 | 62.6 | 59.5 | 62.0 |
| **TAFT + Deeplab V3+** (5-shot, w/ multi-scale) | 59.4 | 69.5 | 64.5 | 60.7 | 63.5 |

Table 4: PASCAL-$5^i$ mean Intersection-over-Union (mIoU) with and without multi-scale input test

| Models | split-0 | split-1 | split-2 | split-3 | mean |
|---|---|---|---|---|---|
| **TAFT + Deeplab V3+** (1-shot, w/o multi-scale) | 68.8 | 72.4 | 68.4 | 66.8 | 69.1 |
| **TAFT + Deeplab V3+** (1-shot, w/ multi-scale) | 69.8 | 73.4 | 68.9 | 67.9 | 70.0 |
| **TAFT + Deeplab V3+** (5-shot, w/o multi-scale) | 72.7 | 79.7 | 74.4 | 75.0 | 75.5 |
| **TAFT + Deeplab V3+** (5-shot, w/ multi-scale) | 74.0 | 81.0 | 76.0 | 76.0 | 76.8 |

Table 5: PASCAL-$5^i$ binary Intersection-over-Union (IoU) with and without multi-scale input test

| Models | split-0 | split-1 | split-2 | split-3 | mean |
|---|---|---|---|---|---|
| **TAFT + Deeplab V3+** (w/ low-level TAFT) | 57.7 | 69.2 | 62.9 | 59.8 | 62.4 |
| **TAFT + Deeplab V3+** | 59.4 | 69.5 | 64.5 | 60.7 | 63.5 |

Table 6: PASCAL-$5^i$ 5-shot mIoU with and without low-level feature transformation

| Models | split-0 | split-1 | split-2 | split-3 | mean |
|---|---|---|---|---|---|
| **TAFT + Deeplab V3+** (w/ low-level TAFT) | 73.4 | 81.2 | 74.9 | 75.5 | 76.3 |
| **TAFT + Deeplab V3+** | 74.0 | 81.0 | 76.0 | 76.0 | 76.8 |

Table 7: PASCAL-$5^i$ 5-shot binary IoU with and without low-level feature transformation

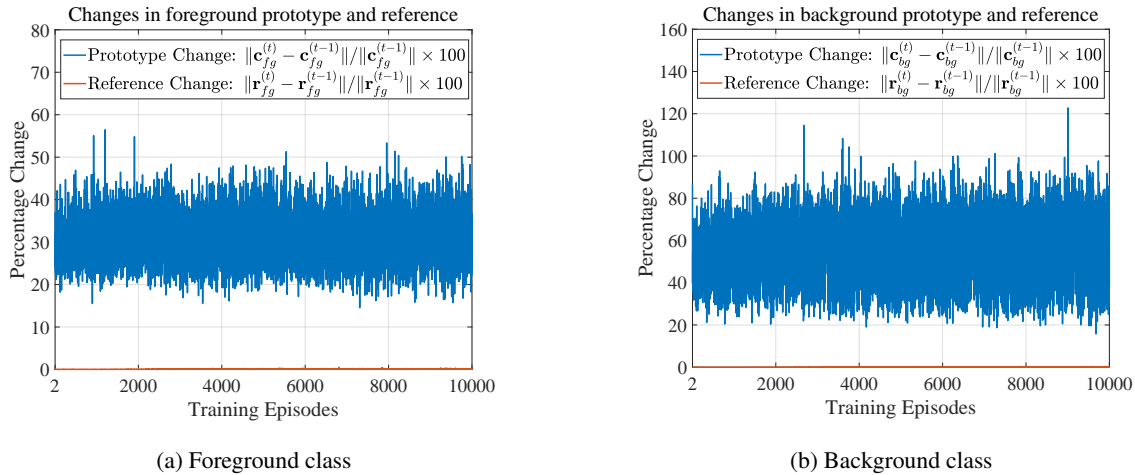

(a) Foreground class    (b) Background class

Figure 2: Comparison of changes in prototypes and references during meta-training

4

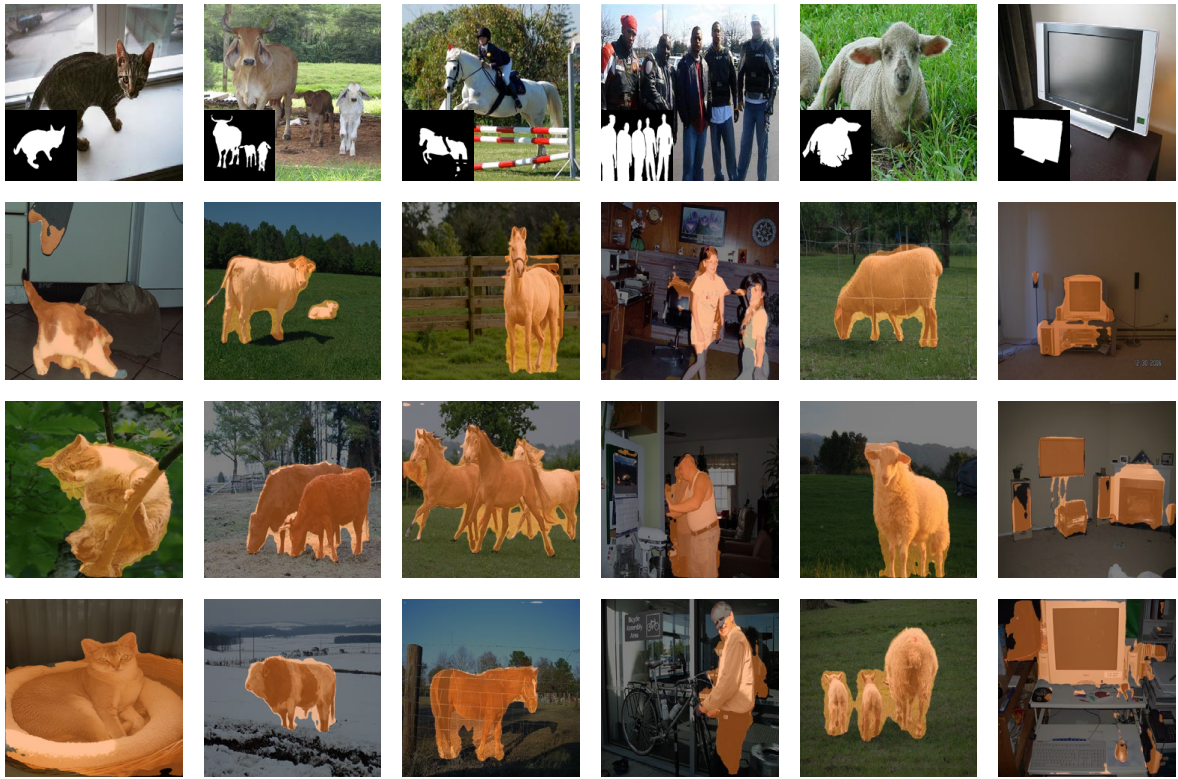

Figure 3: Qualitative results of TAFT + Deeplab V3+ for 1-shot segmentation

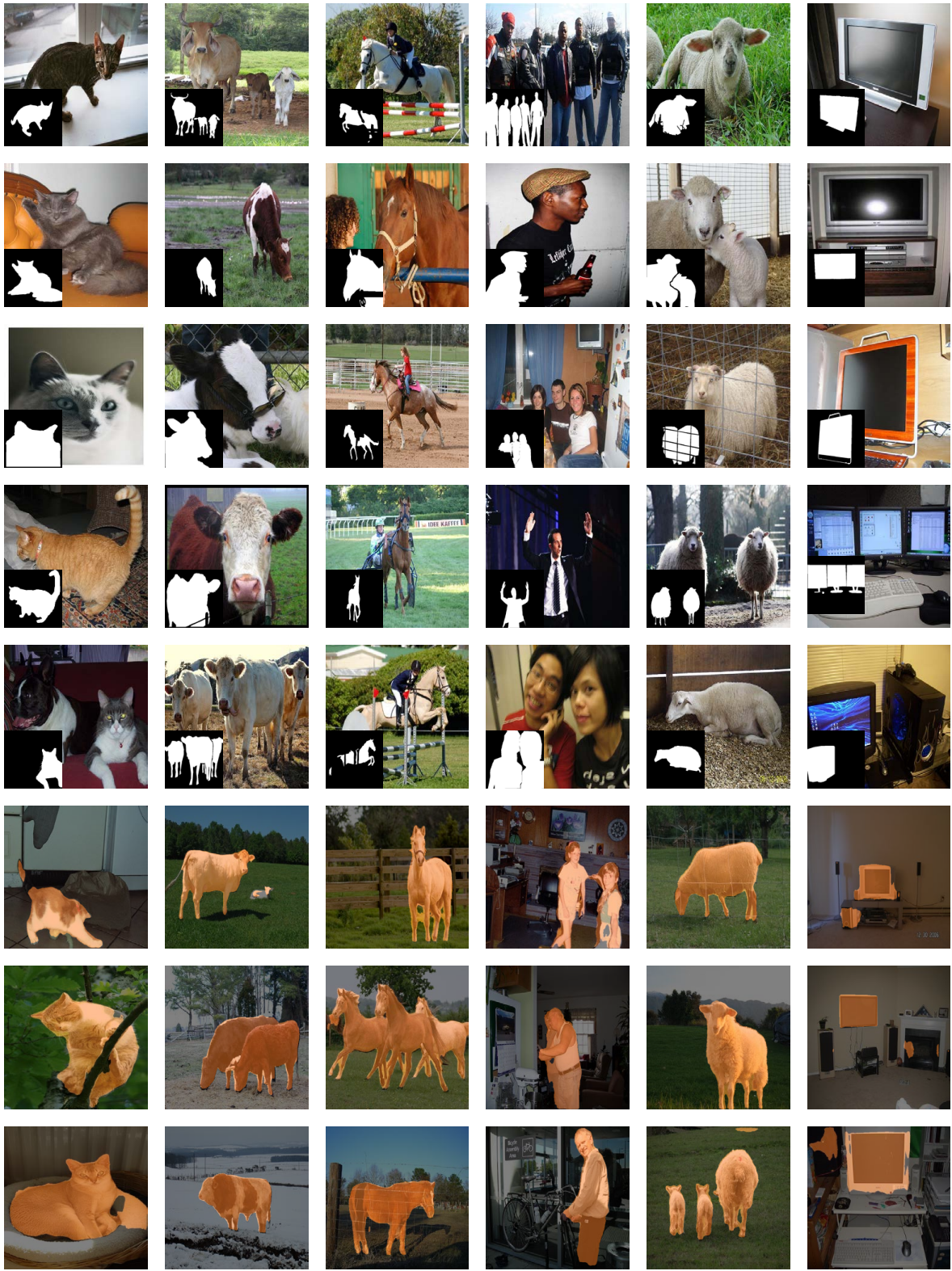

Figure 4: Qualitative results of TAFT + Deeplab V3+ for 5-shot segmentation


\end{document}